\documentclass[11pt]{article}
\usepackage[preprint]{acl}
\usepackage{times}
\usepackage{latexsym}
\usepackage[T1]{fontenc}
\usepackage[utf8]{inputenc}
\usepackage{inconsolata}
\usepackage{graphicx}
\usepackage{xspace,mfirstuc,tabulary}
\usepackage{rotating}
\usepackage{amsmath}
\usepackage{booktabs}
\usepackage{tabularx}
\usepackage{multirow}
\usepackage{makecell}
\usepackage{enumitem}
\usepackage{subcaption}
\usepackage[normalem]{ulem}
\usepackage{xcolor} 
\usepackage{tcolorbox}
\tcbuselibrary{many,breakable}
\newcommand{\ul}[1]{\uline{#1}}

\title{Assisted Counterspeech Writing at the Crossroads of Hate Speech and Misinformation}

\author{
Genoveffa Martone\textsuperscript{1,2} \quad
Helena Bonaldi\textsuperscript{1} \quad
Marco Guerini\textsuperscript{1} \\
\textsuperscript{1}Fondazione Bruno Kessler, Italy,
\textsuperscript{2}Università Cattolica del Sacro Cuore, Italy, \\
\texttt{\{gmartone, hbonaldi, guerini\}@fbk.eu},
}

\definecolor{fc}{RGB}{173, 76, 39}
\definecolor{ngo}{RGB}{46, 120, 160}

\newtcolorbox{ngobox}[1][]{%
  colframe=ngo,
  colback=white,
  boxrule=1pt,
  arc=5pt,
  left=10pt, right=10pt, top=10pt, bottom=10pt,
  enhanced,
  overlay={
    \node[anchor=north west, inner sep=0pt]
      at ([xshift=-3pt,yshift=3pt]frame.north west)
      {\includegraphics[width=5mm]{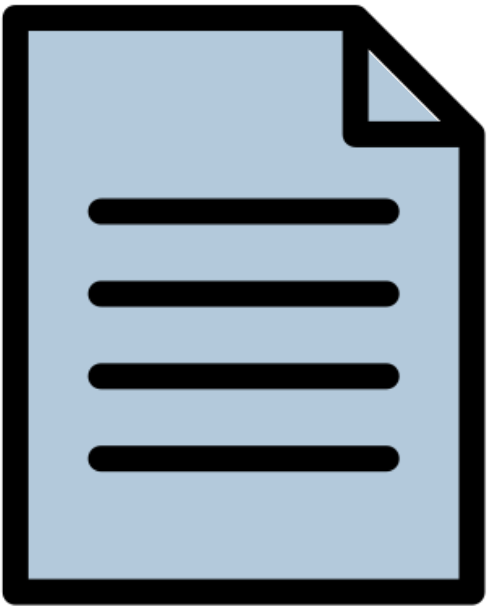}};
  },
  #1
}

\newtcolorbox{fcbox}[1][]{%
  fontupper = \linespread{.85}\fontsize{10pt}{12pt}\selectfont,
  colframe = fc,
  colback  = white,
  segmentation style={solid},
  boxrule=1pt,
  left=10pt,
  right=10pt,
  top=10pt,
  bottom = 10pt,
  middle = 2pt,
  arc=5pt,
  before skip = 10pt,
  after skip = 10pt,
  enhanced,
  overlay={
    \node[anchor=north west, inner sep=0pt]
      at ([xshift=-3pt,yshift=3pt]frame.north west)
      {\includegraphics[width=5mm]{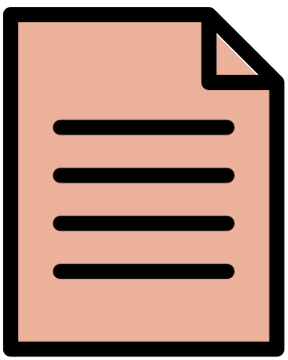}};
  },
  #1
}

\newtcolorbox{fcground}{%
  fontupper = \linespread{.85}\fontsize{10pt}{12pt}\selectfont,
  colframe = fc, 
  colback  = white,
  segmentation style={solid},
  boxrule=1pt,
  left=2pt,
  right=2pt,
  top=2pt,
  bottom = 2pt,
  middle = 2pt,
  arc=5pt,
  before skip = 3pt
}

\newtcolorbox{hs}{%
  fontupper=\linespread{.6}\selectfont,
  after skip=3pt,
  colframe = black,
  colback  = white,
  sharp corners = southwest,
  segmentation style={solid},
  arc=2mm,
  boxrule=1pt,
  left=2pt,
  right=2pt,
  top=2pt,
  bottom = 2pt,
  middle = 2pt
}

\newtcolorbox{csreply}{%
  fontupper=\linespread{.6}\selectfont,
  before skip=3pt,
  after skip = 3pt,
  colframe = black,
  colback  = white,
  sharp corners = southeast,
  segmentation style={solid},
  boxrule=1pt,
  arc=2mm,
  left=2pt,
  right=2pt,
  top=2pt,
  bottom = 2pt,
  middle = 2pt
}

\begin{document}

\maketitle

\begin{abstract}
Hate speech and misinformation frequently co-occur online, amplifying prejudice and polarization. Given their scale, using Large Language Models (LLMs) to assist expert counterspeech (CS) writing has gained interest, yet prior work has addressed these phenomena separately. 
We bridge this gap by studying CS generation in contexts where both hate and misinformation co-occur. 
We test three knowledge-driven generation strategies:
first we prompt an LLM with 
fact-checkers' guidelines and fact-checking articles;
secondly, with NGOs' guidelines and reports;
thirdly, we create a mixed strategy that combines guidelines and documents from both. 23 experts revise the generated CS, which are assessed via human and automatic metrics. While LLMs produce adequate CS in 40\% of cases, expert edits substantially improve naturalness, exhaustiveness, and adherence to guidelines. Based on the post-edited CS, the mixed strategy proves to be the most effective in crowdsourcing evaluation, pairing strong factual correction with stereotype mitigation and empathetic engagement. We release a dataset of
hateful and misinformed claims with expert-verified CS and supporting knowledge.
\end{abstract}

\noindent
\textbf{Warning:} This paper contains unobfuscated examples some readers may find offensive.

\section{Introduction}

Hate speech and misinformation are permeating internet spaces, and 
these phenomena are often intertwined \cite{cazzamatta2025global}: hate speech, aggressive attacks, and partisan rhetoric appear more frequently together with misinformation than with accurate content \cite{hameleers2022civilized}. Additionally, exposure to misinformation can intensify existing prejudices and biases, leading to increased hate towards marginalized groups. This \textit{toxic synergy} intensifies their overall effect, weakening social trust and fueling greater polarization and division \cite{kim2024toxic}. 

In this context, counterspeech (CS) has emerged as a promising strategy to individually tackle both online hate and misinformation. Despite the interconnection of these phenomena, they are mainly addressed separately, by different professionals following distinct guidelines. On the one hand, CS against hate consists of empathetic replies which aim to contrast hateful content via cogent reasons and fact-bound arguments \cite{schieb2016governing}, and it is typically written by NGO operators.
On the other hand, CS against misinformation is usually referred to as \textit{verdicts}, which are short texts written in a non-partisan, non-emotive style, aimed at justifying the truthfulness of a statement \cite{guo2022survey}, and can be produced by fact-checkers \cite{wintersieck2017debating}.
Given the scale of online hate and misinformation, using Large Language Models (LLMs) to automate or assist experts in writing CS against these phenomena has gained growing interest, though prior work has mostly treated them separately \cite{chung2021empowering, mun2024counterspeakers, zeng2024justilm, russo2025face}.
To address this gap, we recognize their interconnection and tackle them jointly, studying to what extent counterspeech against the co-occurrence of hate and misinformation can be automated and which strategy is most effective to approach these phenomena altogether.

\begin{figure}[t!]
    \centering
    \includegraphics[width=0.9\columnwidth]{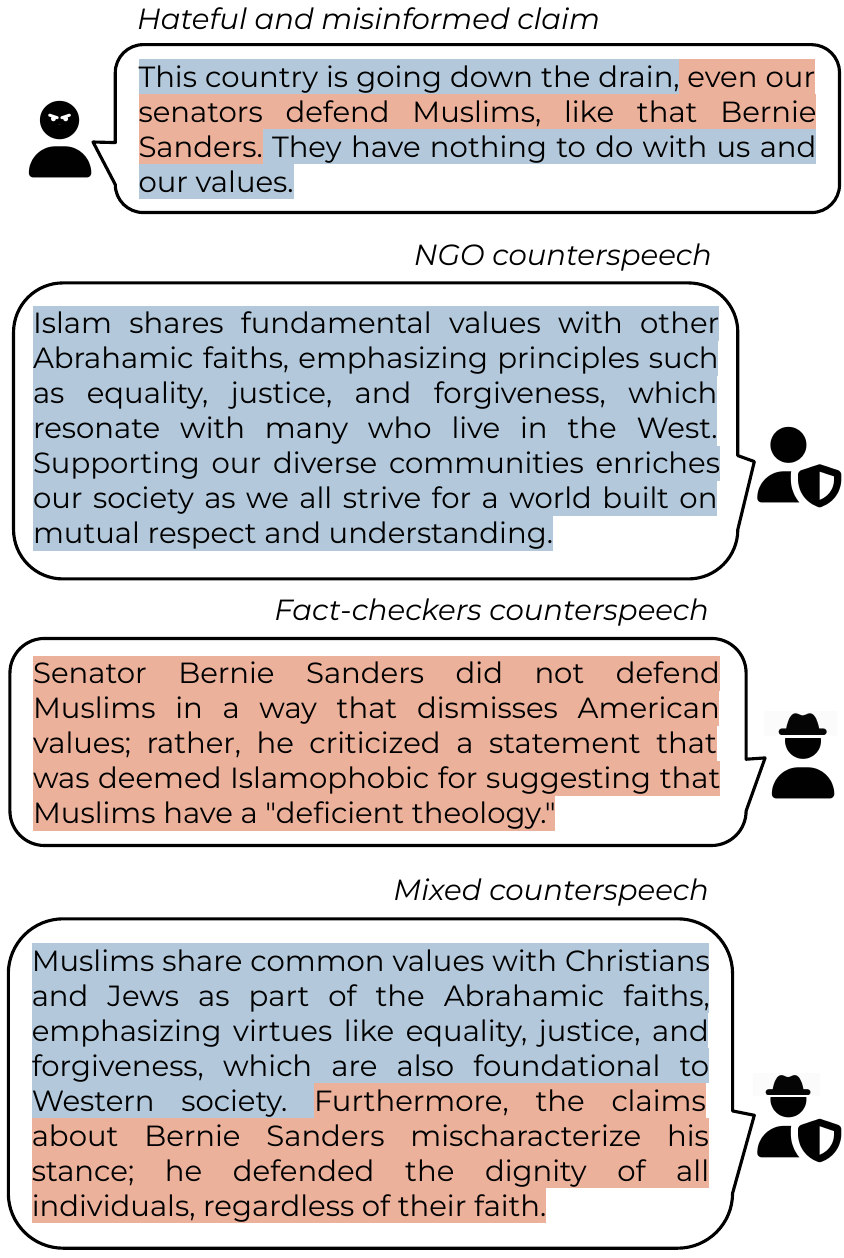}
    \caption{An example of hateful and misinformed claim with CS obtained with the three tested strategies. The text highlighted in orange represents the misinformation-related text, while the light blue refers to the hateful-related text.}
    \label{fig:3strat_example}
\end{figure}

First, we collect expert written guidelines used by NGOs and fact-checkers to respectively counter online hate and misinformation and consider over 2k fact-checking articles and 280 NGO reports to be used as external knowledge. The collected knowledge and guidelines are then given as input to a commercial LLM to produce knowledge-driven CS with three generation configurations:
1) the \textbf{NGO strategy}, following NGO guidelines typically used to counter online hate and employing anti-stereotype reports written by NGOs as external knowledge; 2) the \textbf{fact-checkers strategy}, following the European Code of Standards for Independent Fact-Checking Organizations,\footnote{\url{https://efcsn.com/code-of-standards/}} and using fact-checking articles as external knowledge; 3) the \textbf{mixed strategy},
for which we create a new set of guidelines specifically aimed at addressing the co-occurrence of hate and misinformation, and where both types of documents are used as external knowledge. An example of a hateful claim containing misinformation with CS obtained with the three strategies is shown in Figure \ref{fig:3strat_example}.

\begin{figure*}[t!]
    \centering
    \includegraphics[width=\textwidth]{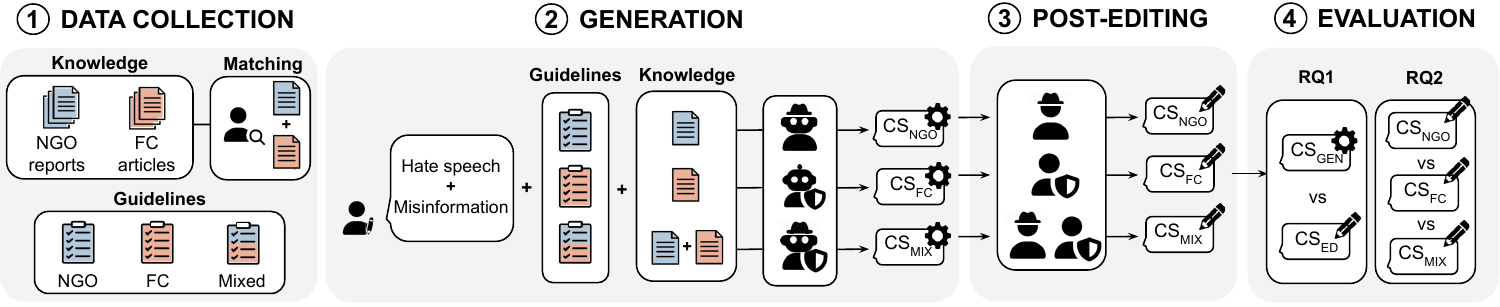}
    \caption{The pipeline of this work: first, the external knowledge and guidelines are collected. Then, they are used as input to produce CS in the three strategies. The generated CS is post-edited by experts and finally evaluated via automatic and human evaluation.}
    \label{fig:pipeline}
\end{figure*}

In our study, we collaborate with 23 NGO operators and fact-checkers from 7 organizations across 4 countries, all expert in CS writing, who revise
the CS generated with the three strategies
by checking the adherence to the respective guidelines, the exhaustiveness of the information provided, and the naturalness of the text. 
In fact, despite the recent major advancements of LLMs, previous studies have shown that they often generate vague CS, mainly denouncing the hateful content without directly engaging with it, and relying on repetitive, unnatural-sounding patterns \cite{tekiroglu2020generating, bonaldi2024is, munbeyond}. Another major concern when deploying such models without human supervision is the risk of producing harmful or inaccurate content \cite{bonaldi-etal-2024-nlp}.
Therefore, we perform both automatic and human evaluations on generated and post-edited pairs to assess their \textbf{quality} via \textbf{RQ1}: 
\textit{To what extent can an LLM assist professionals in CS production against hate and misinformation?}
We find that while annotators are unnecessary in 40\% of cases, their edits improve naturalness and exhaustiveness, align better with expert guidelines, and remove repetitive patterns caused by model stereotyping of fact-checkers and NGO roles.
We then retain the post-edited CS (preferred by both automatic and human metrics) and compare the three strategies through extensive human evaluation to answer \textbf{RQ2}: \textit{Which strategy is most \textbf{effective} in countering the co-occurrence of hate and misinformation?} Our mixed strategy uniquely matches the fact-checkers approach in correcting factual inaccuracies while also mitigating stereotypes and fostering empathy comparably to the NGO approach. An overview of our pipeline is shown in Figure \ref{fig:pipeline}. 
Our dataset, which includes 324 examples of hateful and misinformed statements targeting six marginalized groups, as well as generated and expert post-edited versions of the CS and supporting external knowledge, are available at the following link: \url{https://github.com/LanD-FBK/counterspeech_against_hate_and_misinfo}.

\section{Related work}

\paragraph{CS against hate}

In recent years, there has been a growing interest in automating the production of CS against hate  \cite{bonaldi-etal-2024-nlp}. While part of these studies have relied on CS data available on the web \cite{mathew2019thou, vidgen2020detecting, albanyan-etal-2023-finding} or synthetically obtained \cite{ashida2022towards, vallecillo2023automatic}, others have involved human expertise in the definition of the guidelines and data collection \cite{conan-2019, fanton2021human}. In general, language models have shown recurring problems in generating human-like CS against hate, which can be mitigated by human intervention. In particular, automatically obtained CS tends to be little specific and convincing \cite{tekiroglu2020generating, bonaldi2024is}, generically denouncing the stereotypes presented in the hate speech without directly addressing them \cite{munbeyond}.
Many studies have tried to obtain CS closer in quality to what humans produce, by targeting specific aspects of the generation, such as its personalization \citep{de2021toxicbot, dougancc2023generic, cima2025contextualized} and argumentative cogency \cite{furman-etal-2023-high, bonaldi2024is}. 
Finally, the generation of more informative CS has also been explored, in particular through knowledge-guided generation, where knowledge is first automatically retrieved and then used to generate CS \cite{chung2021towards, jiang2025rezg, russo2025trenteam}.

\paragraph{CS against misinformation}
Also studies focusing on CS generation against misinformation can involve human expertise \cite{russo2023countering, he2023reinforcement}. 
Key evaluation aspects include \textit{readability}, meaning how understandable a CS is; \textit{plausibility}, indicating the persuasiveness of the CS; \textit{faithfulness}, in terms of how accurately the CS reflects the reasoning of the model \citep{guo2022survey}.
While earlier approaches employed attention \cite{popat2018declare} and rule-based techniques \cite{gad2019exfakt}, the currently most employed approaches include summarization \cite{atanasova2020generating, kotonya2020explainable, russo2023countering, russo2023benchmarking}, prompting \cite{russo2025euroverdict}, and Retrieval-Augmented Generation \citep{zeng-gao-2024-justilm, russo2025face}.

\paragraph{CS against hate and misinformation}
While both CS generation against hate and misinformation have been approached by the NLP community as separate tasks, limited research has been dedicated to addressing these two phenomena when they occur at the same time. \citet{lisker2025debunking} focused on the generation of CS against conspiracy theories, only a small portion of which contained hate speech (17 examples, i.e. 11\%). Moreover, even if the authors experimented with a ``fact-based'' CS strategy, they did not provide external knowledge to the language model when generating CS, thus obtaining generic responses often containing made up information. \citet{saha2024integrating} instead, focused on adopting CS strategies against hate to counter misinformation, but did not address their overlap.

\section{Methodology}
This section outlines our methodology. We first collect guidelines and external knowledge 
for model inference, then describe the creation of hateful and misinformed claims, the three CS generation strategies, and the expert annotation process used to revise 
the generations.
Excluding expert annotation, the pre–CS generation phase involved many manual steps: guideline curation, 
external knowledge preparation,
and claim creation, and took approximately 5 person-months of effort.

\subsection{Guidelines collection} \label{subsec:guidelines}
As a first step, we redact the guidelines on how to obtain the CS following the fact-checkers, NGO, and mixed strategies. These serve two purposes: to instruct the model and to provide a uniform reference for expert annotators, reducing potential small variation across their organizations. All guidelines are validated by the annotators before starting with the annotation. The complete guidelines provided to the annotators are reported in Appendix \ref{appendix:annot_guidelines}. 

\paragraph{Fact-checkers guidelines} For what regards the fact-checkers strategy, the guidelines are derived from the 
European Code of Standards for Independent Fact-Checking Organizations.
Specifically, the code emphasizes that fact-checking should focus on verifiable facts and statistics, clearly cite the sources consulted, giving preference to primary over secondary ones, and present the results with precise, non-partisan, non-emotive language.

\paragraph{NGO guidelines} As regards the NGO strategy, we employ the
Get the Trolls Out's project guidelines,\footnote{\url{https://getthetrollsout.org/stoppinghate}}
which highlight the importance of adopting a polite and kind tone, attacking the message rather than the person who wrote it, refraining from using divisive language (e.g. ``fascist'', ``ignorant''), empathizing with fears or anxieties that might underlie the expression of hate, and challenging hurtful stereotypes with alternative viewpoints and evidence. 

\paragraph{Mixed guidelines}  
Fact-checkers and NGO guidelines have some overlapping characteristics. For example, in both cases it is desirable to have polite, fluent and relevant CS \cite{he2023reinforcement, russo2023countering}. At the same time, there are some differences: for instance, fact-checkers' guidelines stress the importance of presenting facts with precise, non-partisan, non-emotive language. On the other hand, when contrasting hate speech, NGO guidelines highlight the importance of empathizing with fears or anxieties that might underlie the expression of hate.
To combine the two approaches, we collaborated with expert annotators to develop a new set of guidelines seeking a balance between these perspectives.
For what regards the style of the intervention, on one hand, the evidence-based approach of fact-checkers is emphasized, highlighting the importance of relying on credible sources. On the other hand, in accordance with NGO recommendations, expressing support for the targeted group is promoted, alongside maintaining a polite and empathetic stance toward the person spreading hate. This is in line with existing literature which shows how, when speaking directly to misinformed individuals, empathetic communication should be used \cite{ecker2022psychological}. For what regards the content of the intervention, the mixed guidelines aim to counter both the negative stereotypes and the factual inaccuracies contained in the claim.

\subsection{External knowledge collection} \label{subsec:knowledge_collection}
In order for the model to properly simulate an NGO operator or a fact-checker's methodology, we collect NGO reports and fact-checking articles to be used as reference material during the generation. 
We focus our data collection on material relevant to one of the following marginalized groups:
Muslims, the LGBTQIA+ community, migrants, women, people with disabilities, and Jews.

\paragraph{Fact-checking articles}
The retrieval of fact-checking articles was performed by crawling Google Fact Check Tools Explorer,\footnote{\url{https://toolbox.google.com/factcheck/explorer/search/list:recent;hl=en}} a search engine for fact-checking articles. First, a list of keywords relevant to the marginalized groups of interest was compiled (the procedure for keywords definition and the complete list of keywords we employed are available in Appendix \ref{appendix:fact_checking_articles}). Then, the Google Fact Check Tools API was used to retrieve the list of relevant URLs resulting from the search performed with each keyword as a query. The actual fact-checking articles were obtained using the \texttt{newspaper4k} library.\footnote{\url{https://pypi.org/project/newspaper4k/}} 
In total, 8,920 fact-checking articles were retrieved. 

Following the indications of the fact-checkers we collaborated with, to ensure a reliable standard for the quality of the fact-checking articles, only the articles written by signatories of the International Fact-Checking Network code of principles\footnote{\url{https://ifcncodeofprinciples.poynter.org/}} were retained.
Subsequently, the collected articles underwent a manual selection aimed at 
identifying content that could be used to spread hate, discrimination, or negative feelings toward specific targets. In particular, an article was selected if (i) it focused on one of the marginalized groups of our interest, rather than on individuals, and either (ii) it contrasted a false claim that could reinforce harmful stereotypes or promote implicit hostility against the target group \citep[e.g., \textit{``The true author of Anne Frank’s diary was an American novelist''} from Full Fact,][]{fullfact2024annefrank},
or (iii) it gave more context about a true claim that could still be twisted or generalized to support negative narratives \citep[e.g., \textit{``In the UK, women earn less than men.''},][]{fullfact2016ukgenderpaygap}.
In total, 2,398 entries were manually assessed, out of which 427 met all the inclusion criteria.

An example of a collected article from \citet{apnews2023canadabillLGBTQ} in Associated Press News follows:

\begin{fcbox}
\small
\noindent 

\textbf{Claim:} A proposed bill in Canada would subject anyone who misgenders others or engages in anti-LGBTQ protests to prosecution and a \$25,000 fine. \\
\\
\textbf{Fact-checking:} False. The bill would not institute a blanket ban on misgendering and anti-LGBTQ protests. The legislation, introduced by members of the opposition party in the Legislative Assembly of Ontario, would allow the province’s attorney general to temporarily prohibit people from engaging in acts of intimidation, such as threats or homophobic protests, within 100 meters (328 feet) of certain properties. [...]
\end{fcbox}

\paragraph{NGO reports}
As a second step, we collected anti-stereotype reports produced by NGOs, addressing false narratives and stereotypes about the groups of our interest. Unlike fact-checking articles, no comprehensive source exists for this type of content. Therefore, we manually searched for appropriate material on websites of organizations committed to dismantling oppressive narratives and promote human rights 
(the full list of the websites' domains we used is available in Appendix \ref{appendix:fact_checking_articles}).
The chosen articles usually highlight a myth about a group, accompanied by an \textit{anti-stereotype} that counters it with evidence, offers alternative perspectives, and explains the misleading origins of the stereotype, as illustrated in this example from \citet{pgaction_myths_v_realities}:

\begin{ngobox}
\small
\noindent
\textbf{Myth}: LGBTI people are requesting “special rights.” \\

\noindent \textbf{Anti-stereotype}: This is not true. There are no special rights being claimed by or for LGBTI people. They are entitled to enjoy the same human rights and fundamental freedoms to which every human being is entitled. Regretfully, these rights and freedoms are denied to millions of people around the world just because of their sexual orientation and gender identity. [...]

\end{ngobox}
\noindent In total, we collected 280 pairs of myth and anti-stereotype, distributed across all target marginalized groups.
\subsection{Hateful messages with misinformation}
Starting from the fact-checking articles, we manually created hateful turns including
the claim verified by the article, as well as a related stereotype.
For example, the hateful and misinformed message created starting from the fact-checking article and NGO report shown as examples in section \ref{subsec:knowledge_collection} is the following: ``\textit{Did you hear that Canada is trying to make it illegal to even protest against LGBTQ stuff? What's next, you can't even express your opinion?}”.
In this way, 213 hateful statements were created.
Then, each of these messages was also matched with all the available anti-stereotypes addressing a similar misconception to that conveyed by the hateful message. This matching was first performed automatically, by computing the sentence-level semantic similarity\footnote{The similarity was obtained with the \texttt{all-mpnet-base-v2} sentence transformers model.} between each hateful turn and each myth present in the NGO reports. Then, following a preliminary review of the outputs, only matching pairs with a similarity higher than 0.4 were manually checked by two of the authors.
After this manual revision, the NGO knowledge to be used for each hateful message is obtained by concatenating all the relevant myth–anti-stereotype pairs identified.
In this way, 108 triplets of hateful messages containing misinformation (from hereafter, just ``claims''), corresponding NGO reports, and fact-checking articles were retained.

\subsection{CS generation}

We experiment with three generation configurations: the NGO strategy (CS$_{NGO}$), the fact-checkers strategy (CS$_{FC}$), and the mixed strategy 
(CS$_{MIX}$). The first two aim at emulating, respectively, an NGO operator's and a fact-checker's work, while the last is a combination of the two. 

In all three cases we employ GPT-4o mini \cite{openai2024gpt4omini} in a zero-shot setting, with varying instructions and external knowledge as reference. In particular, for each configuration we employ a simplified version of the strategy-specific guidelines described in Section \ref{subsec:guidelines}, together with the collected external knowledge, i.e., either a fact-checking article for CS$_{FC}$, an NGO report for CS$_{NGO}$, or both for  CS$_{MIX}$. The hyperparameters as well as the complete prompts used to instruct the model are available in Appendix \ref{appendix:generation_details}. For each claim, one CS was generated with each of the three configurations, thus obtaining 324 pairs in total.

\subsection{Expert annotation} \label{subsec:annotation}
For revising the generated CS, we collaborated with 23 experts in CS writing (11 fact-checkers and 12 NGO operators). 
First, a preliminary meeting with the annotators was set up to present the task and the First-AID platform,
which is specifically designed for annotating dialogues based on external knowledge \cite{menini2025first}.
Each participant was assigned to the strategy related to their professional expertise: therefore, NGO operators focused on the generations obtained with the NGO approach, whereas fact-checkers focused on the fact-checkers configuration. Additionally, both groups were also required to work on the mixed strategy. 
In all cases, participants were presented with a subset of claims and generated CS pairs (each pair was annotated only once).
Their task consisted in modifying the CS,
ensuring naturalness and a consistent tone with the specific strategy's guidelines used (as described in Section \ref{subsec:guidelines}), and verifying the accuracy and relevance of the information in the CS with respect to the provided document(s). Moreover, participants were also asked to annotate which parts of the document(s) contained the information included in the CS (i.e., the ``ground text''), and they could leave comments if needed. 
An example of generated (CS$_{GEN}$) and post-edited CS (CS$_{ED}$) using the fact-checkers strategy, along with the annotated ground text, is shown below.

\begin{hs}
\small \noindent \textbf{Claim}: We can't just let people pour into the country without any checks. It's unfair, even Obama said it!
\end{hs}

\begin{csreply}

\small \noindent \textbf{CS$_{GEN}$}: While Barack Obama did express concerns about immigration and the need for stronger borders, \textit{he also emphasized that merely securing the borders is not a comprehensive solution to immigration management.} It's important to recognize that successful immigration policies involve a balance of security and the recognition of the contributions that immigrants make to society. \end{csreply}

\begin{csreply}
\small \noindent \textbf{CS$_{ED}$}: \uline{When Senator of Illinois, in 2005,} Barack Obama did express concerns about immigration and the need for stronger borders. \textit{However, he also said that merely securing the borders does not solve immigration management. }
\end{csreply}

\begin{fcground}
\small \noindent \textbf{Ground text 1}: True. The quote comes from a 2005 news conference where Obama, then a U.S. Senator from Illinois, spoke about immigration and those who employ immigrants. \\
\textbf{Ground text 2}: A longer version of the clip shows Obama saying, “securing the borders alone does not solve immigration management.
\end{fcground}

The underlined modifications add context, while that in italics is rephrased to improve readability. The final sentence is removed because it's incoherent with fact-checkers' guidelines.

\section{RQ1: Quality}
We perform both automatic and human evaluation to
answer RQ1: \textit{To what extent can an LLM assist professionals in CS production against hate and misinformation?}

\subsection{Automatic evaluation}
We 
assess the impact of human post-editing and the quality of the CS in terms of repetitiveness, readability, lexical and syntactic metrics.
\paragraph{HTER} The \textit{Human-targeted Translation Edit Rate} (HTER) assesses the post-editing effort
in terms of word insertion, deletion, modification or word sequences shifts
\citep{snover2006study}. 
It ranges from 0 to 1, with higher values indicating more extensive edits. 
A score higher than 0.4 indicates modifications requiring a comparable effort to complete rewriting \cite{turchi2013coping}. 

\paragraph{Repetition Rate} It measures the lexical diversity of a corpus as the proportion of non-singleton n-grams it contains \citep{bertoldi2013cache}. The RR is agnostic of the corpus size as it is computed as the geometric mean obtained on a sliding window.
Consistently with other works, we calculate it as the average obtained on five random shuffles of a corpus \cite{bonaldi2024is}.

\paragraph{Readability} We evaluate readability using three metrics. The \textit{Flesch Reading Ease Score} (FRES) measures text difficulty based on sentence and word length, with higher scores indicating easier reading \cite{flesch1948new, kincaid1975derivation}.
The \textit{Flesch-Kincaid Grade Level} (FKG) expresses readability in U.S. school grade levels, where lower values indicate easier text \cite{kincaid1975derivation}. Finally, we consider the \textit{proportion of complex words} (CW), defined as words with three or more syllables \cite{gunning1952technique}.

\paragraph{Lexical and syntactic analysis} We also perform a lexical analysis and compute the 
\textit{most added and removed words} by the annotators, and the \textit{length} of the 
provided ground text. For what regards syntactic analysis, we computed the \textit{average} and \textit{maximum syntactic depth} (ASD, MSD) and the \textit{number of sentences} (NST).

\begin{table}[h]
\small
\centering
\resizebox{\columnwidth}{!}{%
\begin{tabular}{llcccc}
\toprule
\textbf{Config.}& \textbf{Ann.}& \textbf{n}&  \textbf{p$_{mod}$} &\textbf{HTER}&\textbf{HTER$_{m}$} \\
\midrule
\textbf{CS$_{FC}$}  & FC              & 108        & \underline{45.4\%} & \underline{0.215} & 0.474                          \\
\textbf{CS$_{NGO}$} & NGO             & 108        & 73.1\%                          & 0.264                          & 0.360                          \\
\textbf{CS$_{MIX}$} & FC              & 32         & \textbf{43.8\%}                 & \textbf{0.128}                 & \textbf{0.292}                 \\
\textbf{CS$_{MIX}$} & NGO             & 76         & 68.4\%                          & 0.241                          & \underline{0.352} \\ \bottomrule
\end{tabular}}
\caption{
\textit{Config} is the generation strategy, \textit{Ann.} the annotator's profession, \textit{n} the number of examples, p$_{mod}$ the percentage of modified examples, HTER and HTER$_{m}$ show the average HTER over all pairs and modified pairs only, respectively.}
\label{tab:hter_results}
\end{table}

\subsection{Automatic evaluation results}

\begin{table*}[h]
\small
\centering
\begin{tabular}{llllcccccc}
\toprule
\textbf{Config.}     & \textbf{Ann.} & \textbf{RR$_{gen}$}& \textbf{RR$_{ed}$}  & \textbf{FRES$_{gen}$} & \textbf{FRES$_{ed}$} & \textbf{FKG$_{gen}$} & \textbf{FKG$_{ed}$} & \textbf{CW$_{gen}$} & \textbf{CW$_{ed}$} \\ \midrule
\textbf{CS$_{FC}$}  & FC &  \textbf{3.030}& \textbf{2.640}  & \textbf{24.565}& \textbf{29.416}& \textbf{17.046}& \textbf{15.943}& \textbf{0.222}      & \textbf{0.21}      \\
\textbf{CS$_{NGO}$} & NGO    &  7.269 & 5.497  & {\ul 17.989}& 21.817& {\ul 17.644}& {\ul 16.455}& 0.26                & 0.25               \\
\textbf{CS$_{MIX}$} & FC   &  \underline{4.764}& 4.876  & 15.209& 21.717& 19.251& 17.577& {\ul 0.249}         & {\ul 0.236}        \\
\textbf{CS$_{MIX}$} & NGO    &  5.830 & \underline{4.603}  & 13.481& {\ul 22.599}& 19.677& 16.948& 0.253               & 0.241              \\ \bottomrule
\end{tabular}
\caption{Repetitiveness and readability results}
\label{tab:rr_read_results}
\end{table*}

\paragraph{Post-editing effort} In total, experts took 44 hours to complete all annotations, with 8 minutes per dialogue on average. The HTER results are shown in Table \ref{tab:hter_results}. For all configurations the average HTER is below 0.4, showing a general good quality of the generations.
If we consider the results according to the type of annotator, fact-checkers are those with the lowest average HTER and percentage of modified pairs (p$_{mod}$), both in the CS$_{FC}$ and CS$_{MIX}$ configurations. Even if we consider HTER on modified pairs only (HTER$_{m}$), results are below 0.4 for all configurations except for CS$_{FC}$. This shows that, although for this configuration annotators intervened less often on the generations, when they did their edits were substantial. 
NGO operators, in contrast, intervened more frequently than fact-checkers (73.1\% and 68.4\% of modified examples for CS$_{NGO}$ and CS$_{MIX}$, respectively), but their HTER is always lower than 0.4, even for HTER$_{m}$.
Additionally, both fact-checkers and NGO operators showed a decreased effort on CS$_{MIX}$ respectively compared to CS$_{FC}$ and CS$_{NGO}$, both in terms of frequency and intensity of interventions. This suggests a higher quality of the CS produced with CS$_{MIX}$.

\paragraph{Linguistic complexity} Turning to the \textbf{repetitiveness} (Table \ref{tab:rr_read_results}), by comparing the RR$_{gen}$ and RR$_{ed}$, it is possible to notice how human intervention makes the generations less repetitive in all cases, except for the fact-checkers' intervention on the CS$_{MIX}$ configuration. However, the delta is the smallest among the various configurations (0.112) and might be explained by the low number of examples for this combination. 
On the other hand, the highest and second highest deltas are obtained by NGOs in CS$_{NGO}$ and CS$_{MIX}$, respectively, showing a higher reduction of the repetitiveness in the generations. In fact, the CS obtained with the CS$_{NGO}$ strategy showed the highest RR$_{gen}$ to start with: this difference is related to the structure of the external knowledge used. In particular, while fact-checking articles are unique configuration-wise, each pair of myth and anti-stereotype can be repeated across multiple NGO reports in the same configuration.

\begin{table*}[h]
\small
\centering
\begin{tabular}{llcccccc}
\toprule
\textbf{Config.}     & \textbf{Ann.} & \textbf{MSD$_{gen}$} & \textbf{MSD$_{ed}$} & \textbf{ASD$_{gen}$} & \textbf{ASD$_{ed}$} & \textbf{NST$_{gen}$} & \textbf{NST$_{ed}$} \\ \midrule
\textbf{CS$_{FC}$}  & FC              & {\ul 9.138}          & {\ul 8.953}         & \textbf{7.768}       & 7.526               & 2.0                  & 2.138               \\
\textbf{CS$_{NGO}$} & NGO             & \textbf{8.870}       & \textbf{8.268}      & {\ul 7.796}          & \textbf{6.985}      & 2.0                  & {\ul 2.481}         \\
\textbf{CS$_{MIX}$} & FC              & 9.812                & 9.625               & 8.453                & 8.166               & 2.0                  & 2.250               \\
\textbf{CS$_{MIX}$} & NGO             & 9.342                & 9.092               & 8.210                & {\ul 7.473}         & 2.0                  & \textbf{2.684}      \\
\bottomrule
\end{tabular}
\caption{Syntactic metrics results}
\label{tab:syntactic_results}
\end{table*}

Human post-editing also allows to obtain higher \textbf{readability} scores for all configurations, as the FRES increases, and both FKG and CW decrease, as shown in Table \ref{tab:rr_read_results}. In particular, the CS$_{FC}$ configuration is the most readable, both before and after human post-editing. Moreover, for both FRES and FKG, although the CS$_{MIX}$ configuration generates the least readable text, it is also the setting in which post-editing brings the highest improvement. 
The higher readability scores in the post-edited data can also be explained by the results obtained on the \textbf{syntactic metrics} (Table \ref{tab:syntactic_results}).
In particular, for all configurations, both the average and maximum syntactic depth decreased with post-editing and the number of sentences increased, showing that annotators tended to split long and complex sentences into clearer and shorter ones. In fact, both FRES and FKG assume that shorter words and phrases are easier to understand. 

\paragraph{Lexical analysis} 
To understand the content of human annotations, 
we examine the most added and removed n-grams, which reveal recurring patterns across the three configurations.
In particular, for CS$_{FC}$, the two most removed words are \textit{claim} and \textit{false}, possibly underlying 
a high frequency of the model in generating sentences structured
as \textit{the claim [...] is false}, thus stereotyping the fact-checkers communicative style. A similar stereotyping occurs for CS$_{NGO}$, where the most removed words are \textit{important} and \textit{recognize}, which represent also the most common bigram in the generations with this configuration. The tendency of GPT-4 o-mini to often generate periphrases such as \textit{it's important to recognize}, as well as \textit{it's important to understand}
(as shown by the high reduction of \textit{understand} too), is coherent with similar findings from past work on zero-shot CS generation,
where this tendency appeared across different models, including Mistral \cite{bonaldi2024is} and earlier versions of the GPT family, such as GPT-3 and GPT-3.5 \cite{russo2023countering}.

Finally, also for CS$_{MIX}$ the phrase \textit{it’s important to recognize} appeared frequently in the model’s output. NGO operators generally reduced its use during post-editing, whereas fact-checkers retained this formula at a higher rate even after revision. This different behavior can be related to the lower frequency of this pattern in CS$_{FC}$ generations than in CS$_{NGO}$'s, thus making fact-checkers less sensitive to such repetitions.

\paragraph{External knowledge} Finally, we look into the type of external knowledge (ground text) considered in the CS$_{MIX}$ strategy.\footnote{Before performing this analysis, three examples were removed as outliers, more detailed are present in Appendix \ref{appendix:automatic_metrics}} As described in Section \ref{subsec:annotation}, adding the ground text was part of experts' annotation, thus information about it is available only for the post-edited CS. 
In the CS$_{MIX}$ configuration, regardless of the annotator type, the percentage of ground text coming from fact-checking articles is prevailing: it represents 60.6\% of the ground text for examples annotated by fact-checkers and 54.94\% for examples annotated by NGO operators. A qualitative analysis showed that in the majority of cases annotators tended to insert additional content taken from the fact-checking article rather than from the NGO report. This possibly indicates an insufficient reliance of the model on this type of knowledge, although all generations except three examples refer to content from both types of articles.

\subsection{Human evaluation}
We recruit 6 graduate-level volunteer annotators with expertise in evaluating machine-generated content to assess pairs of generated (CS$_{GEN}$) and post-edited CS (CS$_{ED}$) replying to the same claim.
We select all pairs with an HTER value of 0.39 or higher. This threshold is chosen for three main reasons: (i) based on a manual inspection, to ensure that annotators assess pairs with a sufficient degree of modification to be clearly perceptible; (ii) it is close to the 0.4 threshold identified by \citet{turchi2013coping}; and (iii) it provides a fair balance across generation strategies, yielding 24 examples from CS$_{NGO}$ and 20 from CS$_{FC}$ and CS$_{MIX}$. In total, 78 pairs are evaluated, with 21 of them (seven per strategy) assessed twice to compute inter-annotator agreement.
The selected 0.39 threshold aims to maximize the coverage of annotated examples while keeping the task cognitively manageable: even with this strict filtering, annotators reported that the options were often very similar, making comparisons challenging. Lowering the threshold would have increased the number of evaluated pairs but also their similarity, significantly increasing the difficulty of the task.

For each pair to be evaluated, respondents are presented with the specific guidelines with which the CS are obtained, the claim, and the generated and post-edited CS in random order, marked as ``CS \#1'' and ``CS \#2''. 
Respondents are asked to rank the two presented CS options along three dimensions, defined based on the main shortcomings that currently limit the full automation of CS production in real-world settings.
First, language models can fail to adhere to CS writing guidelines and generate inaccurate or harmful content, which may amplify existing harm instead of countering it \cite{bonaldi-etal-2024-nlp}. Second, they often produce vague and generic CS that mainly condemns hateful language without directly engaging with it, frequently relying on repetitive or unnatural patterns \cite{tekiroglu2020generating, bonaldi2024is, munbeyond}.
To capture these phenomena, we ask respondents to assess the two CS options by 
selecting the CS which (i) aligns more closely with the provided guidelines, (ii) feels more natural and easier to understand, and (iii) provides more precise and exhaustive information.
To prevent cross-contamination, each survey uses a between-subjects design and includes examples generated with a single strategy, allowing annotators to apply consistent guidelines throughout.

\subsection{Human evaluation results}
Table \ref{tab:survey1_results} shows the results of the human evaluation in terms of percentage preference of the respondents for CS$_{GEN}$ and CS$_{ED}$, grouped by evaluated dimension.\footnote{As the total number of evaluations per dimension is 99, percentage and raw preferences correspond.} For each dimension, CS$_{ED}$ is significantly preferred over CS$_{GEN}$: significance is calculated with a one-sided binomial test.\footnote{One star indicates a p-value $<$ 0.01, while two stars a p-value $<$ 0.001.}
The overall Cohen's kappa, calculated as the average pairwise agreement between annotators, is 0.21, while the percentage agreement is 0.57. These results are in line with subjective tasks \cite{bonaldi2024is}, and can be interpreted as there is an overall preference for CS$_{ED}$, although annotators can disagree on instance-level judgments. More details are in Appendix \ref{appendix:iaa}. \\

\noindent Taken together, the results of the automatic and human evaluation allow us to answer \textbf{RQ1}: 
\textit{To what extent can an LLM assist professionals in CS production against hate and misinformation?}
Specifically, the automatic evaluation shows that the model generates good quality CS for all configurations, which overall do not require excessive post-editing, with an average HTER < 0.4. In particular, CS$_{MIX}$ required less post-editing from both fact-checkers and NGO operators than CS$_{FC}$ and CS$_{NGO}$, respectively. However, human intervention is still necessary as it makes generations less repetitive, more readable and less syntactically complex. It also allows to remove the stereotyped expressions that are frequently generated by the model in all configurations. Finally, according to the survey's results, annotators play a fundamental role in making generated CS more natural, exhaustive, and adherent to guidelines. 

\begin{table}
\small
    \centering
    \begin{tabular}{lllcc}
    \toprule
       \textbf{Dimension}  & \textbf{CS$_{GEN}$} & \textbf{CS$_{ED}$} \\ \midrule
        Guidelines & 31 & \textbf{68}**  \\
        Exhaustiveness & 35 & \textbf{64}* \\
        Naturalness &  36 & \textbf{63}*\\
        \bottomrule
    \end{tabular}
    \caption{Results of the survey assessing annotators' impact in terms of percentage preferences for CS$_{GEN}$ and CS$_{ED}$.}
    \label{tab:survey1_results}
\end{table}

\section{RQ2: Effectiveness}
After evaluating generation quality, we consider all the post-edited CS, as they
were preferred by automatic and human assessments, to conduct an extensive human study and determine which strategy (NGO, fact-checker, or mixed) is most effective at countering hate and misinformation.

\subsection{Human evaluation}
The survey was first tested in an internal evaluation pilot with a small group of participants and then administered on Prolific (more details
are available in Appendix \ref{appendix:human_eval}). First, the study's guidelines and the dimensions to evaluate are presented: 
we ask how good the presented CS is in (i) challenging the factual accuracy of the claim (FACT), (ii) challenging the stereotype presented in the claim (STER), (iii) promoting empathy for the targeted group (EMP), and (iv) discouraging others from agreeing with the claim (DISC).
Participants are asked to answer from the perspective of a bystander, choosing among one of the following options: ``No attempt made'', ``Very Poor'', ``Poor'', ``Good'', ``Very Good''.
Before the main survey, they need to pass two comprehension checks to ensure their understanding of the task. These consist of two unambiguous cases in which the CS clearly does not address either the factual inaccuracy or the stereotype contained in the claim.
Failing a comprehension check twice causes the exclusion of the participant from the survey. Two attention checks are also randomly inserted among the main survey's questions.

In total, 37 participants were involved, evaluating 32 different surveys containing the entirety of the examples (5 surveys are annotated twice to calculate inter-annotator agreement, corresponding to 50 pairs). Beyond the comprehension and attention checks, each survey included on average 10 claim-CS pairs to be evaluated, balanced according to (i) the LLM configuration used to obtain the CS (ii) the annotator type (fact-checker or NGO operator), (iii) the group targeted in the claim; 
(iv) the external knowledge used as reference.
Pairs were randomized to avoid any order effect.

\begin{table}[t!]
\small
\centering
\begin{tabular}{lllll}
\toprule
\textbf{Config.} & \textbf{\makecell[l]{FACT}} & \textbf{\makecell[l]{STER}} & \textbf{\makecell[l]{DISC}} & \textbf{\makecell[l]{EMP}} \\
\midrule
\textbf{CS$_{FC}$} & \underline{3.370}** & 2.565 & \underline{2.880} & 1.963 \\
\textbf{CS$_{NGO}$} & 2.429 & \underline{2.938}* & 2.714 & \textbf{2.964}** \\
\textbf{CS$_{MIX}$} & \textbf{3.458}** & \textbf{3.224}** & \textbf{3.131}** & \underline{2.907}** \\
\bottomrule
\end{tabular}
\caption{Mean ratings of CS strategies across the four evaluation dimensions.}
\label{tab:he_results}
\end{table}

\subsection{Results}

The results of the human evaluation are available in Table \ref{tab:he_results} and allow us to answer RQ2, i.e., \textit{What is the most effective strategy to contrast hate and misinformation when occurring together?}\footnote{Four surveys were excluded from the analyses: more details are present in Appendix \ref{appendix:iaa}.}
The stars in the table indicate a
statistically significant difference with respect to
the lowest-performing configuration column-wise,
computed with the Mann-Whitney U test. 
By looking at the results, it is possible to notice how CS$_{MIX}$ obtained the best results on almost all dimensions, except for \textit{Promoting Empathy}, where it is the second best.
In particular, for \textit{Factual accuracy} it performs comparably well to that of CS$_{FC}$: both are significantly better than CS$_{NGO}$. At the same time, CS$_{MIX}$ is also as good as CS$_{NGO}$ at challenging the stereotype presented in the claim, as both are significantly better than CS$_{FC}$, and the same applies to \textit{Promoting Empathy}. Finally, for what regards \textit{Discouraging Agreement}, CS$_{MIX}$ is the configuration obtaining the best results, which are significantly higher than those obtained by CS$_{NGO}$.
Overall, these results show how the mixed strategy is the most successful in obtaining a performance as good as the configuration specialized in each aspect (i.e., CS$_{FC}$ for factual accuracy, CS$_{NGO}$ for challenging stereotype and fostering empathy), but this does not affect the results on the other aspects. 
In fact, while both CS$_{FC}$ and CS$_{NGO}$ satisfy one or more of the evaluated aspects, they do so at the cost of other dimensions: this is confirmed by the comments left from the surveys' respondents (see Appendix \ref{appendix:human_eval} for more details). On the other hand, CS$_{MIX}$ is the only strategy that successfully addresses all the evaluated aspects at the same time, thus representing the most suitable approach to contrast the co-occurrence of hate speech and misinformation. 
The overall Cohen's kappa and Spearman's rho are 0.25 and 0.38, respectively. We hypothesize that low agreement reflects annotators interpreting the scale differently. By collapsing it into positive (``Good'' and ``Very Good'') and non-positive categories, kappa raises to 0.29 and rho to 0.42, confirming that part of the initial low agreement stemmed from scale nuances 
(see Appendix \ref{appendix:iaa} for more details).

To answer \textbf{RQ2}, we can conclude that CS$_{MIX}$ is the most effective in tackling the co-occurrence of hate and misinformation, as it uniquely balances the ability to correct factual inaccuracies of CS$_{FC}$, and it fosters empathy and challenges the implicit hostility comparably to CS$_{NGO}$. It also significantly reduces
agreement with the hateful claim more than CS$_{FC}$.

\section{Conclusion}
We investigated three knowledge-driven CS generation strategies against the co-occurrence of hate and misinformation: the fact-checkers strategy, the NGO strategy, and a mixed strategy combining the two. Generated content was revised by expert NGO operators and fact-checkers, and evaluated using human and automatic metrics to address two research questions. First, we asked to what extent CS generation can be automated. While all strategies produced high-quality outputs, human intervention reduced repetitiveness, improved readability, and removed stereotypical expressions. Human evaluations confirmed that post-edited CS is more adherent to guidelines, exhaustive, and natural than raw generations. Second, we assessed which strategy best counters hate and misinformation. Results show the mixed strategy is the only one capable of effectively challenging the factual accuracy of claims, performing comparably to the fact-checkers strategy, while also addressing stereotypes and fostering empathy toward the targeted group, comparably to the NGO strategy. Moreover, it significantly reduces agreement with the hateful claim more than the fact-checkers approach. Overall, the mixed strategy is the most effective for addressing both hate and misinformation, highlighting the potential of human-assisted CS generation for this dual challenge.

\section*{Acknowledgements}
This work was partially supported by the European Union’s CERV fund under grant agreement No. 101143249 (HATEDEMICS). We are grateful to the following NGOs, fact-checking organizations and all annotators for their help: ALDA (Association Europeenne Pour La Democratie Locale), FUNDEA (Fundacion Euroarabe De Altos Estudios), MALDITA (Fundacion Maldita.Es Contra la Desinformacion: Periodismo educacion Investigacion y Datos En Nuevos Formatos), CENTRA (Fundacion Pública Andaluza Centro De Estudios Andaluces M.P.), CESIE ETS (CESIE ETS), TFCF (The Fact-Checking Factory S.r.l.), SOS MALTA (Solidarity And Overseas Service Malta), VSA (Victim Support Agency), CEO (Fundacja Centrum Edukacji Obywatelskiej), DEMAGOG (Stowarzyszenie Demagog), NASK (Naukowa I Akademicka Siec Komputerowa - Panstwowy Instytut Badawczy).

\bibliography{anthology, custom}

\clearpage
\appendix

\section{Appendix}

\subsection{Guidelines for the annotators}  \label{appendix:annot_guidelines}
\begin{figure*}
    \centering
    \includegraphics[width=\textwidth]{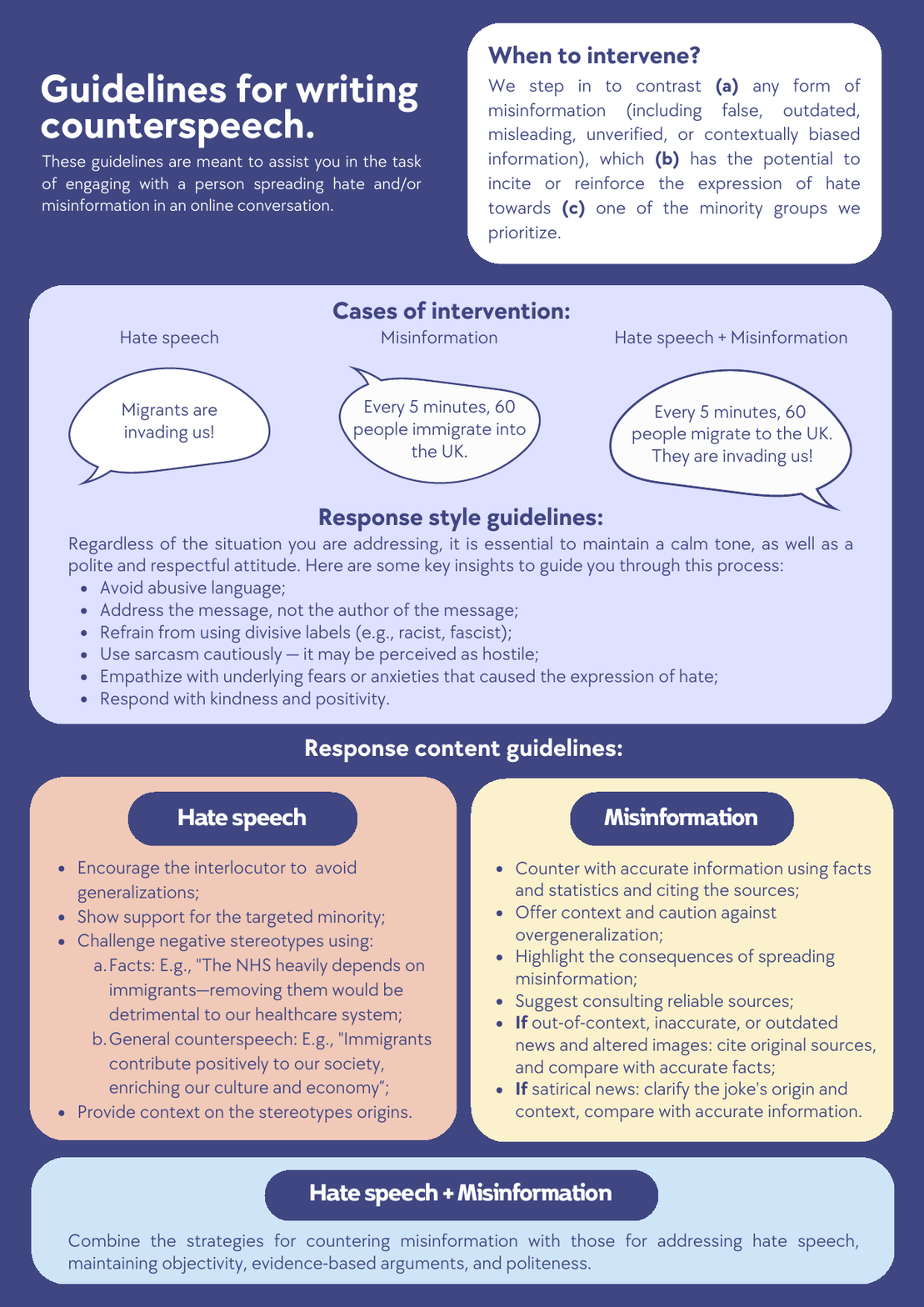}
    \caption{Guidelines for the annotation of the CS generated with the mixed configuration.}
    \label{fig:guidelines}
\end{figure*}

The complete guidelines given as a reference to annotators are shown in Figure \ref{fig:guidelines}: fact-checkers and NGO operators were told that for the strategies matching their roles they could refer to their own professional standards.
The mixed guidelines were consulted by both groups and represent 
a compromise between the fact-checkers and NGO approach. Annotators were also asked to ensure that both the misinformation and the negative stereotype were addressed in the CS generated with this strategy.

\subsection{Documents collection} \label{appendix:fact_checking_articles}
\paragraph{Fact-checking articles} The selection process of the keywords to use for querying Google Fact Check Tools was informed by a review of the terms used in previous studies that focused on contrasting hate speech against the same marginalized groups considered in this work \citep{bonaldi2023weigh, ousidhoum2019multilingual, waseem2016hateful, poletto2017hate}. 
As a starting point, the keyword lists included both identity terms and prejudice terms. Identity terms are words that are closely related to the identity of a group, whereas prejudice terms reflect stereotypical or biased expressions commonly attributed to that group \citep{bonaldi2023weigh}. 
The keywords lists were expanded to also include common slurs used against the groups in question, as well as popular hashtags used in conversations surrounding them. To this aim, we also consulted 
dedicated Wikipedia pages\footnote{\href{https://en.wikipedia.org/wiki/Category:Antisemitic_slurs}{Antisemitic slurs}, 
\href{https://en.wikipedia.org/wiki/Category:Islam-related_slurs}{Islam-related slurs}, 
\href{https://en.wikipedia.org/wiki/Category:LGBT-related_slurs}{LGBT-related slurs}, 
\href{https://en.wikipedia.org/wiki/Category:Pejorative_terms_for_people_with_disabilities}{Pejorative terms for people with disabilities}, 
\href{https://en.wikipedia.org/wiki/Category:Pejorative_terms_for_women}{Pejorative terms for women}.} featuring derogatory terms directed against 
the groups of or interest \citep{ross2017measuring, waseem2016hateful}. 

Following this preliminary collection, the keyword lists were manually reviewed and refined, excluding terms that were deemed too niche, outdated, or unlikely to yield relevant results when submitted to Google Fact Check Tools. The complete keywords lists resulting from this process are available in Table \ref{tab:appendix_keywords}:

\begin{table*}
\small
    \centering
    \begin{tabularx}{\textwidth}{lX}
    \toprule
        \textbf{Targeted group} & \textbf{Keywords}\\
        \midrule
        Muslims & muslim, islam, terrorist, jihadi, jihad, ragheadterror, arab, koran,
quran, sharia, towel head, rag head\\
        LGBTQIA+ & gay, homosexual, homosexuality, lgbt, lgbt+, lgbti, lgbtq+, lgbtq,
faggot, gender, lesbian, trans, transgender, transsexual, queer, sexual, sex, heterosexual, dyke, gay pride\\
       Migrants  & migrant, immigrant, refugee, immigration, foreigner, migration, foreign, rapefugees, invasion, invade, refugeesnotwelcome\\
       Women  & woman, feminism, feminist, gender, female, harassment, feminazi, shithole, cunt, blameonenotall, notallmen, victimcard, sexual assault, victim card\\
       People with disabilities  & disabled, disability, autistic, blind, deaf, retard, downies, downy, paralympics, wheelchair\\
        Jews & jew, jewish, holocaust, judaism, nazi, nazism, genocide\\
         \bottomrule
    \end{tabularx}
    \caption{Keywords used to query Google Fact Check Tools for retrieving fact-checking articles related to the groups of our interest.}
    \label{tab:appendix_keywords}
\end{table*}

The collection of the articles was conducted in English, and it leveraged the Google Fact Check Tools Application Programming Interface (API) in combination with the identified keywords. For each category, a Python script iterated through the list of keywords, issuing a query to the Google Fact Check Tools APIs. Each query retrieved up to 1000 fact-checking articles. Additionally, we also scraped the entire FullFact\footnote{\url{https://fullfact.org/}} website.

\paragraph{NGO reports} The URL domains from which we extracted the employed NGO reports are the following: \url{https://aasas.ca}, \url{https://communitiesinc.org.uk}, \url{https://cpdonline.co.uk}, \url{https://developmenteducation.ie}, \url{https://edmo.eu}, \url{https://efcl.org}, \url{https://healthjournalism.org}, \url{https://iine.org}, \url{https://jrseurope.org}, \url{https://medium.com}, \url{https://rapecrisis.org.uk}, \url{https://safeandequal.org.au}, \url{https://simpl4all.eu}, \url{https://worldrelief.org}, \url{https://www.adl.org}, \url{https://www.bpas.org}, \url{https://www.coolmindshk.com}, \url{https://www.enar-eu.org}, \url{https://www.etf.europa.eu}, \url{https://www.infomigrants.net}, \url{https://www.learningforjustice.org}, \url{https://www.nata.org}, \url{https://www.osce.org}, \url{https://www.pgaction.org}, \url{https://www.psychologytoday.com}, \url{https://www.rescue.org}, \url{https://www.strongfamilyalliance.org}, \url{https://www.thearcbc.org}, \url{https://www.unh.edu}, \url{https://www.vera.org}, \url{https://www.weforum.org}.

\subsection{Generation details} \label{appendix:generation_details}
We employ \texttt{gpt-4o-mini-2024-07-18}. Generations are obtained with maximum token limit of 100 and a temperature of 0.8.
Below, the prompts employed for generating CS with the three configurations are shown: \\

\noindent \textbf{CS$_{NGO}$}: ``You are an NGO operator who provides counterstatements to negative stereotypes contained in hateful messages.
Given the hateful claim provided in HATER\_TURN, generate a two-sentence-long CS following the guidelines provided in NGO\_GUIDELINES and using the innformation contained in the NGO\_ARTICLE. 
Answer by referring exclusively to NGO\_ARTICLE, don't cite the sources in brackets. \\

\noindent HATER\_TURN: \textless HATER\_TURN\textgreater \\
\noindent NGO\_GUIDELINES: \textless NGO\_GUIDELINES\textgreater \\
\noindent NGO\_ARTICLE: \textless NGO\_ARTICLE\textgreater'' \\

\noindent \textbf{CS$_{FC}$}: ``You are a fact-checker who provides counterstatements to misinformation contained in hateful messages.
Given the hateful claim provided in HATER\_TURN, generate a two-sentence-long CS following the guidelines provided in FC\_GUIDELINES and using the information contained in the FC\_ARTICLE.
Answer by referring exclusively to FC\_ARTICLE, don't cite the sources in brackets. \\

\noindent  HATER\_TURN: \textless HATER\_TURN\textgreater \\
\noindent  FC\_GUIDELINES: \textless FC\_GUIDELINES\textgreater \\
\noindent  FC\_ARTICLE: \textless FC\_ARTICLE\textgreater'', \\

\noindent \textbf{CS$_{MIX}$}: ``You are a counterspeaker who provides counterstatements to hateful messages.
Given the hateful claim provided in HATER\_TURN, generate a two-sentence-long CS following the guidelines provided in MIX\_GUIDELINES.
You must necessarily use the facts contained in the FC\_ARTICLE to contrast misinformation and the content from the NGO\_ART-ICLE to contrast stereotypes.
Answer by referring exclusively and equally to FC\_ARTICLE and NGO\_ARTICLE, don't cite the sources in brackets. \\

\noindent HATER\_TURN: \textless HATER\_TURN\textgreater \\
\noindent MIX\_GUIDELINES: \textless MIX\_GUIDELINES\textgreater\\
\noindent NGO\_ARTICLE: \textless NGO\_ARTICLE \textgreater\\
\noindent FC\_ARTICLE: \textless FC\_ARTICLE \textgreater''\\

Additionally, the guidelines provided for the three configurations are the following: \\

\noindent \textbf{FC\_GUIDELINES}: Counteract misinformation with accurate and verifiable facts and statistics.
Provide evidence for every factual statement made in the counter speech.
Mention the sources on which the counter speech is based. 
Keep the counter speech impartial and avoid political partisanship.
Formulate counter speech using a precise, factual, and non-emotive language. \\

\noindent \textbf{NGO\_GUIDELINES}: Avoid abusive language.
Challenge the claim, not the person who wrote it.
Refrain from using divisive labels (e.g., racist, fascist).
Express support to those who might be under attack.
Counter hate with kindness, positivity, mutual respect, and politeness.
Empathize with underlying fears or anxieties that caused the expression of hate.
Challenge negative stereotypes using facts and providing context. \\

\noindent \textbf{MIX\_GUIDELINES}: Avoid abusive language and divisive labels (e.g., racist, fascist). 
Challenge the claim, not the person who wrote it.
Counter misinformation with accurate facts, evidence, impartiality, and reliable sources.
Provide context for the misinformed hateful claim.
Express support for those under attack and respond with kindness and respect.
Empathize with underlying fears or anxieties that caused the expression of hate.
Challenge negative stereotypes using facts and providing context. 

\subsection{Outliers removal} \label{appendix:automatic_metrics}
When performing the ground text analysis, three annotated examples were identified as outliers.
One example was discarded because the ground text was only two words long, while another was disproportionately long (almost 5,000 words).
A third case was removed because the annotator 
attached a fact-checking article as external source to a CS generated with the NGO strategy.

\subsection{Details on Survey Evaluating Effectiveness} \label{appendix:human_eval}

\paragraph{First Survey Draft}
The first version of the survey 
included ten pairs of hateful and misleading claims each, along with their corresponding CS$_{ES}$ messages. For each pair, participants were asked to evaluate the 
CS according to the following four dimensions:

\begin{itemize}
    \item \textit{The CS challenged the factual accuracy of the claim;}
    \item \textit{The CS challenged the stereotype presented in the claim;}
    \item \textit{The CS made me feel empathy for the targeted group;
    \item The CS would discourage others from agreeing with the claim.}
\end{itemize}

After reading each claim-CS pair, participants were asked to indicate their level of agreement with these statements on a five-point Likert scale, including the following options:  ``Strongly disagree'', ``Somewhat disagree'', ``Neither agree nor disagree'', ``Somewhat agree'', ``Strongly agree''. Respondents had also the option to leave comments. 

Before getting to the main survey, the participants were provided with an introduction to the task. This section specifically explained the purpose of the study, offered a brief definition of CS, and presented the four evaluation dimensions.
Finally, a disclaimer was included to inform the respondents about the sensitive content of the study.

\paragraph{Pilot Testing}

To test the first survey draft, a small pilot was conducted with a group of five volunteer respondents who had not received any prior information about the aim of the study in advance. The participants included 4 PhD students and a researcher, and they were all familiar with designing surveys for NLP research.

The aim of this test was to evaluate the clarity of the task and of the instructions provided, to assess the suitability of the response scale, and to obtain a preliminary approximation of the time required to complete the full survey. On average, the five annotators finished their questionnaires in about 15 minutes, and after that, they were given the opportunity to provide feedback.

The respondents shared several doubts about the survey. The first concerned the response scale used: annotators found that the lower end of the Likert scale was confusing, particularly when interpreting the ``Strongly disagree'' option. It was unclear whether selecting it meant that the CS did not address a given dimension at all (e.g. it did not challenge the factual accuracy of the claim), 
or if it did address that dimension, but in a bad way (e.g. it provided confusing statistics to challenge the factual accuracy of the claim).
Similarly, the ``Neither agree nor disagree'' option was perceived as misleading, as participants could not understand whether it represented partial effectiveness (being positioned in the middle of the scale) or complete irrelevance, as implied by its neutrality. 

Further confusion emerged from the CS messages generated with the CS$_{NGO}$ configurations, which by design failed to address the factuality of the claim. In such cases, the participants incorrectly concluded that the absence of evidence in the CS to challenge the claim was due to the fact that the claim was true and did not contain misinformation. However, as already mentioned, all claims contained both a negative stereotype and a false piece of information.

A final issue concerned the perspective to adopt when answering the questions. In fact, while three out of four items were phrased in an impersonal way that clearly required respondents to evaluate the CS from a bystander's point of view (i.e., \textit{``The CS challenged/would discourage [...]''}), one was formulated in the first person (i.e., \textit{``The CS made me feel [...]''}). This inconsistency wrongly suggested that the evaluation should be based on the respondent's own perspective. 

\paragraph{Final Survey}
The full text of the guidelines included in the final survey are presented in Table \ref{tab:survey_guidelines}, while comprehension checks and attention checks are presented in Table \ref{tab:comprehension_checks} (for both comprehension checks the correct answer was \textit{No attempt made}). For the attention checks, the questions were rephrased and asked \textit{``Based on the text above, how good were you asked to say the CS is at [...]?''}, with each of the dimension of interest. 

Participants were recruited through Prolific, and  
as eligibility criteria 
we selected participants who (i) are native English speakers
(ii) have a 100\% approval rate in previous Prolific studies; (iii) have completed between 500 and 10,000 prior submissions on the platform. 
Respondents were not allowed to participate to more than one survey in this study. The compensation was set at £7.60 per hour,
in line with the platform's fair payment guidelines.\footnote{\url{https://researcher-help.prolific.com/en/article/2273bd}} 

During data collection, 30 participants were automatically filtered out for not passing the comprehension checks. Therefore, they autonomously returned their submissions, without needing to be paid, allowing other eligible users to take part in the survey.
The average completion time for the surveys was 17 minutes and 27 seconds, with a standard deviation of 10 minutes and 26 seconds.

\begin{table*}
\small
    \centering
    \begin{tabularx}{\textwidth}{X}
    \toprule
      \textbf{Guidelines}  \\
      \midrule
      This research aims to evaluate \textbf{different CS strategies} in response to claims that contain misinformation and stereotypes about marginalized groups.

\textbf{CS} generally refers to direct and polite responses to hateful or misinformed claims, written with the intention to challenge their hostile or misleading content and to de-escalate the conversation.

In this task you will be asked to read different types of CS in reply to specific claims, that \textbf{ALWAYS} contain both misinformation AND negative stereotypes. You will then be asked to reply to the following questions:

\begin{itemize}
    \item \textbf{``How good is the CS in challenging the factual accuracy of the claim?''}: this aims to understand how good the CS is in correcting the wrong information contained in the claim.
    \item \textbf{``How good is the CS in challenging the stereotype presented in the claim?''}: this aims to understand the ability of the CS to contrast the negative stereotypes contained in the claim.
    \item \textbf{``How good is the CS in promoting empathy for the targeted group?''}: this aims to understand the ability of the CS to foster empathy towards the group attacked by the claim.
    \item \textbf{``How good is the CS in discouraging others from agreeing with the claim?''}: this aims to assess how likely a reader would be to reject the claim after reading the CS. To do this, consider the perspective of someone reading the exchange online without a clear opinion or prior knowledge of the topic.
\end{itemize}

The possible answers are: \textbf{Very good, Good, Poor, Very poor, }and\textbf{ No attempt made}. Please note that ``No attempt made'' refers to cases where the CS does not address the specific aspect mentioned in the question at all (e.g., it makes no attempt to challenge the presented stereotype).

If you want to leave a comment, you can use the optional comment space under each example.

\textbf{PLEASE NOTE} that this survey contains unobfuscated examples of hate speech that might be offensive to the respondents. \\
  
        \bottomrule
    \end{tabularx}
    \caption{Final survey guidelines}
    \label{tab:survey_guidelines}
\end{table*}

\begin{table*}
\small
    \centering
    \begin{tabularx}{\textwidth}{X}
    \toprule
  \textbf{Comprehension Checks} \\
  \midrule
  \begin{enumerate}
      \item Based on the instructions, how good is the following CS in challenging the factual accuracy of the claim?

\textbf{CLAIM:} \textit{There were more than 100 terrorist attacks made by Muslims in the US in the last year.}

\textbf{CS}: \textit{Islam is a religion of peace rather than violence.}
\item Based on the instructions, how good is the following CS in challenging the stereotype presented in the claim?

\textbf{CLAIM}: \textit{Women should stay at home raising children!}

\textbf{CS}: \textit{I understand your concern, I also feel the same from time to time.}
  \end{enumerate}
\\
\midrule
\textbf{Attention Checks} \\
\midrule
\begin{enumerate}
    \item 
Please, read these two messages carefully:

\textbf{CLAIM:} \textit{Please, answer the following questions selecting ``No attempt made''.}

\textbf{CS}: \textit{This is an attention check. Please take a moment to respond to the following questions, and for each one, kindly select the option labeled ``No attempt made'' as your answer.}
\item 
Please, read these two messages carefully:

\textbf{CLAIM}: \textit{Please, answer the following questions selecting ``Very Good''.}

\textbf{CS}: \textit{This is an attention check. Please take a moment to respond to the following questions, and for each one, kindly select the option labeled ``Very Good'' as your answer.}
\end{enumerate} \\

\bottomrule
    \end{tabularx}
    \caption{Comprehension and attention checks}
    \label{tab:comprehension_checks}
\end{table*}

\subsection{Qualitative analysis of survey 2 comments}
Below are shown some comments regarding CS$_{FC}$, highlighting its ability to generate CS challenging the factual accuracy of the claim but not addressing the implied negative stereotype or raising empathy: \\

\textit{``While this CS did a good job of correcting the incorrect statement [...], it didn't really tackle the stereotype[...]''}  
\begin{flushright}
-- \textit{P\_6,  Prolific survey, CS$_{FC}$}
\end{flushright}

\textit{``This CS does a solid job at fact-checking and debunking misinformation, which is critical in reducing panic or false beliefs about migration numbers. However, it misses the chance to address the xenophobic tone of the claim and to encourage empathy or a more welcoming perspective toward migrants.''}  
\begin{flushright}
-- \textit{P\_1,  Prolific survey, CS$_{FC}$} 
\end{flushright} 
\vspace{1em}

\noindent And for what regards CS$_{NGO}$, the opposite is true: \\

\textit{``Good at challenging the stereotype and promoting empathy. Doesn’t directly refute the claim’s facts.''}  
\begin{flushright}
-- \textit{P\_7,  Prolific survey, CS$_{NGO}$}
\end{flushright}

\textit{``The CS is strong in tone and intent, especially in confronting harmful stereotypes and defending the humanity of trans and LGBTI individuals.''}  
\begin{flushright}
-- \textit{P\_1,  Prolific survey, CS$_{NGO}$}
\end{flushright}
\vspace{1em}

\subsection{Inter-annotator agreement} \label{appendix:iaa}

\paragraph{Survey 1}
Table \ref{tab:survey1_iaa_results} shows the Cohen's kappa agreement results calculated grouped according to the tested dimensions.
The low results for Cohen's kappa and percentage agreement (\% AGR) is common to subjective tasks. Overall, these results can be interpreted as there is an overall preference for post-edited counterspeech, although annotators disagree on instance-level judgments. In particular, the \textit{Naturalness} dimension is the most subjective. A post-hoc interview with the annotators highlighted how, despite selecting only the pairs with highest HTER, they often perceived the two options as very similar. One annotator specifically mentioned that Naturalness was the most difficult dimension to evaluate, as both options were highly natural.

\begin{table}
\small
    \centering
    \begin{tabular}{lllcc}
    \toprule
       \textbf{Dimension}  & $\kappa$ & \% AGR\\ \midrule
        Guidelines & 0.22 & 62 \\
        Exhaustiveness  &  0.25 & 62\\
        Naturalness & 0.00 & 48\\
        \bottomrule
    \end{tabular}
    \caption{Results of survey 1, assessing annotators' impact in terms of number of Cohen's kappa, and percentage agreement.}
    \label{tab:survey1_iaa_results}
\end{table}

\paragraph{Survey 2} 
\begin{table}[]
\begin{tabular}{lcccc}
\toprule
\textbf{Dimension} & \textbf{$\kappa$} & \textbf{$\rho$} & \multicolumn{1}{c}{\textbf{$\kappa_{confl
}$}} & \multicolumn{1}{c}{\textbf{$\rho_{confl}$}} \\ \midrule
\textbf{FACT}      & 0.35              & 0.40            & 0.32                                  & 0.33                                \\
\textbf{STER}      & 0.14              & 0.18            & 0.43                                  & 0.50                                \\
\textbf{EMP}       & 0.19              & 0.20            & 0.30                                  & 0.39                                \\
\textbf{DISC}      & 0.19              & 0.31            & 0.35                                  & 0.44                                \\ \bottomrule
\end{tabular}
\caption{Cohen's Kappa and Spearman's Rho results on the original and on the conflated (0, 1, 2) scale.}
\label{tab:iaa}
\end{table}

Four surveys were excluded from the analyses, as their respondents only used one or two of the available scores: this did not affect the rankings nor the statistical significance of the results, but it reduced the sample for calculating inter-annotator agreement to 80 examples.
Table \ref{tab:iaa} shows the Inter-Annotator Agreement results of the survey calculated using Cohen's Kappa and Spearman's Rho.
The results indicate generally low agreement among annotators. The highest agreement corresponds to the factual accuracy correction, with $\kappa$ = 0.348 indicating a fair agreement, and $\rho$ = 0.400 reflecting a moderate correlation. This shows how annotators were reasonably aligned in identifying factual corrections in the CS, even if they did not always assign the same scores. 

We hypothesize that these relatively low results stem from differing annotator interpretations of the scale: in particular, annotators may have treated all negatively or positively connoted options as equivalent. Therefore, we merge the scale into two broader categories: positive (``Good'' and ``Very Good'') and non-positive (``No attempt made'', ``Poor'' and ``Very Poor'').  $\kappa_{confl}$ and $\rho_{confl}$ in
table \ref{tab:iaa} shows that by merging the higher and lower values of the original scale, the inter-annotator agreement is higher in nearly all dimensions, except for the factual accuracy correction.

\end{document}